\documentclass[conference]{IEEEtran}
\usepackage{blindtext, graphicx}
\usepackage{listings}
\usepackage{graphicx}
\usepackage[font=small]{caption}

\usepackage[utf8]{inputenc}

\title{RBPGAN: Recurrent Back-Projection GAN for Video Super Resolution }

\author{
    \IEEEauthorblockN{Marwah Sulaiman\IEEEauthorrefmark{1}, Zahraa Shehabeldin\IEEEauthorrefmark{1}, Israa Fahmy\IEEEauthorrefmark{1}, Mohammed Barakat\IEEEauthorrefmark{1}, Mohammed El-Naggar\IEEEauthorrefmark{1}, \\Dareen Hussein\IEEEauthorrefmark{1}, Moustafa Youssef\IEEEauthorrefmark{1}, Hesham M. Eraqi\IEEEauthorrefmark{2}\textsuperscript{1}}
    \IEEEauthorblockA{\IEEEauthorrefmark{1}Department of Computer Science and Engineering, The American University in Cairo, Cairo, Egypt}
    \IEEEauthorblockA{\{marwahisham, zahraaagamal, israafahmy, mohamedyasser36, Mohamed\_elnaggar,\\ dareenhussein, moustafa.youssef\}@aucegypt.edu}
    \IEEEauthorblockA{\IEEEauthorrefmark{2}Amazon, Last Mile Geospatial Science, Seattle, WA}
    \IEEEauthorblockA{heraqi@amazon.com}
}

\begin{document}

\maketitle
\thispagestyle{empty}
\pagestyle{empty}

\footnotetext[1]{The work was conducted prior to Hesham Eraqi joining Amazon}

\begin{abstract}
Recently, Video Super Resolution (VSR) has become a very impactful task in the area of Computer Vision due to its various applications. In this paper, we propose Recurrent Back-Projection Generative Adversarial Network (RBPGAN) for VSR in an attempt to generate temporally coherent solutions while preserving spatial details. RBPGAN integrates two state-of-the-art models to get the best in both worlds without compromising the accuracy of produced video. The generator of the model is inspired by RBPN system [Haris et al. 2019], while the discriminator is inspired by TecoGAN [chu et al. 2018]. We also utilize Ping-Pong loss [Chu et al. 2018] to increase temporal consistency over time. Our contribution together results in a model that outperforms earlier work in terms of temporally consistent details, as we will demonstrate qualitatively and quantitatively using different datasets.


\end{abstract}

\section{Introduction}

Video Super Resolution (VSR) is the process of generating High Resolution (HR) Videos from Low Resolution (LR) Videos. Videos are of the most common types of media shared in our everyday life. From entertainment purposes like movies to security purposes like security camera videos, videos have become very important. As a result, VSR has also become important. The need to modernize old videos or enhance security camera videos to identify faces became significant over the last years. VSR aims to enhance the quality of videos for these needs.
\bigbreak
Similar and older than VSR is ISR (Image Super Resolution), which is the process of generating a single high-resolution image from a single low-resolution image. Since a video is understood to be a sequence of frames (images), Video Super Resolution can be seen as Image Super Resolution (ISR) applied to each frame in the video. While this is useful because many of the ISR techniques can be slightly modified to apply to VSR, however, there are major differences between VSR and ISR. The main difference is the temporal dimension in videos that does not exist in images. The relationship between a frame in a video and other frames in the video is the reason why VSR is more complex than ISR.
\bigbreak
In this research, various VSR methods will be explored. The methods are mainly clustered into two clusters, methods with alignment and methods without alignment. We will compare between the different methods across different datasets and discuss the results. Out of the methods we studied, we chose 2 models to be the base models for our research paper. We further explore these base models and experiment with them.
\bigbreak

This paper aims to minimize the trade-off between temporal coherence and quality of VSR. To achieve this, we propose a Generative Adversarial Network (GAN) that combines components from each one of the base models to achieve the best of both worlds. Our methodology, experimentation and results are mentioned in this paper respectively. Finally, we conclude the paper in the last section and propose some future recommendations for further

\section{Related Work}

Based on our review of the literature, the Deep Learning based methods that target Video Super Resolution problem can be divided into 2 main categories: methods with alignment, and methods without alignment. Alignment basically means that the input LR video frames should be aligned first before feeding them into the model. Under the methods with alignment, existing models can be divided into two sub-categories: methods with Motion Estimation and Motion Compensation (MEMC), and methods with Deformable Convolution (DC). Under methods without alignment, existing models can be divided into 4 sub-categories: 2D convolution, 3D convolution, RNNs, and Non-Local based. In this section, the state-of-the-art methods belonging to every category will be discussed.

\subsection{Methods with Alignment}
\subsubsection{Motion Estimation and Motion Compensation (MEMC)}\hfill 

First, the Temporally Coherent Generative Adversarial Network (TecoGAN) [Chu et al. 2018] The network proposes a temporal adversarial learning method for a recurrent training approach that can solve problems like Video Super Resolution, maintaining the temporal coherence and consistency of the video without losing any spatial details, and without resulting in any artifacts or features that arbitrarily appear and disappear over time. The TecoGAN model is tested on different datasets, including the widely used Vid4, and it is compared to the state-of-the-arts ENet, FRVSR, DUF, RBPN, and EDVR. TecoGAN has significantly less trainable weights than RBPN and EDVR. It scores PSNR of 25.57, and its processing time per frame is 41.92 ms. TecoGAN is able to generate improved and realistic details in both down-sampled and captured images.However, one limitation of the model is that it can lead to temporally coherent yet sub-optimal details in certain cases such as under-resolved faces and text. \\
Second, the recurrent back-projection network (RBPN) [Haris et al. 2019]. This architecture mainly consists of one feature extraction module, a projection module, and a reconstruction module. The recurrent encoder-decoder module integrates spatial and temporal context from continuous videos. This architecture represents the estimated inter-frame motion with respect to the target rather than explicitly aligning frames. This method is inspired by back-projection for MISR, which iteratively calculates residual images as reconstruction error between a target image and a set of its corresponding images. These residual blocks get projected back to the target image to improve its resolution. This solution integrated SISR and MISR in a unified VSR framework as SISR iteratively extracted various feature maps representing the details of a target frame while the MISR were used to get a set of feature maps from other frames. This approach reported extensive experiments to evaluate the VSR and used the different datasets with different specs to conduct detailed evaluation of strength and weaknesses for example it used the Vid4, and SPMCS which lack significant motions. It proposes an evaluation protocol for video SR which allows to differentiate performance of VSR based on the magnitude of motion in the input videos. It proposes a new video super-resolution benchmark allowing evaluation at a large scale and considering videos in different motion regimes.\\

\subsubsection{Deformable Convolution methods (DC)} \hfill 

The Enhanced Deformable Video Restoration (EDVR) [Wang et al., Wang] model was the winning solution of all four tracks of the NTIRE19 competition. In addition, outperformed the second-best solution. Also, this solution performed better when compared to some of the state-of-the-art solutions. EDVR is a framework that performs different video super-resolution and restoration tasks. The architecture of EDVR is composed of two main modules known Pyramid, Cascading, and Deformable convolutions (PCD) and Temporal and Spatial Attention (TSA). EDVR was trained on the REDS dataset, which contains 240 training videos and 60 videos divided equally for validation and testing. Each video in the REDS dataset is a 100 consecutive frame short clip.

\subsection{Methods without alignment}
\subsubsection{2D convolution} \hfill 

Generative adversarial networks and perceptual losses for video super-resolution [Lucas et al. 2019]. The model uses a GAN to generate high-resolution videos. The generator and the discriminator in the GAN consist both of many convolutional layers and blocks. The generator first generates a high-resolution frame, and the discriminator decides whether the output from the generator is a generated frame or a ground-truth (GT) image. If the discriminator decides it is a generated frame, then the generator uses the output of the discriminator to generate a better, closer to GT, high-resolution frame. The process is then repeated multiple times until the discriminator accepts the output of the generator as a GT image.\\

\subsubsection{3D convolution} \hfill 

The dynamic filter network can generate filters that take specific inputs and generate corresponding features. The dynamic upsampling filters (DUF) [Jo et al. 2018] use a dynamic filter network to achieve VSR. The structure of the dynamic upsampling filter and the spatio-temporal information learned from the 3D convolution led to a comprehensive knowledge of the relations between the frames. DUF performs filtering and upsampling operations and uses a network to enhance the high-frequency details of the super-resolution result.

\subsubsection{RCNNS} \hfill 

RCNNS is a very powerful network [Dieng et al. 2018] developed a stochastic temporal convolutional network (STCN) by incorporating a hierarchy of stochastic latent variables into TCNs, allowing them to learn representations over a wide range of timescales. The network is divided into three modules: spatial, temporal, and reconstruction. The spatial module is in charge of extracting features from a series of LR frames. Temporal module is a bidirectional multi-scale convoluted version Motion estimation of LSTM that is used to extract temporal correlation between frames. The latent random variables in STCN are organized in accordance with the temporal hierarchy of the TCN blocks, effectively spreading them across several time frames. As a result, they generated a new auto-regressive model that combines the computational advantages of convolutional architectures with the expressiveness of hierarchical stochastic latent spaces. The model in STCN is meant to encode and convey information across its hierarchy.

\subsubsection{Non-Local methods} \hfill 

There is a  progressive fusion network for vSR that is meant to make greater use of spatio-temporal information that has shown to be more efficient and effective than existing direct fusion, slow fusion, and 3D convolution techniques through a technique known as Progressive Fusion Video Super-Resolution Networks in Exploiting Non-Local Spatio-Temporal Correlations (PFNL). This is presented in Progressive Fusion Video Super-Resolution Network via Exploiting Non-Local Spatio-Temporal Correlations [Yi et al. 2019]. That enhanced the non-local operation in this progressive fusion framework to circumvent the MEMC methods used in prior VSR techniques.This was done by adding a succession of progressive fusion residual blocks (PFRBs). The suggested PFRB is designed to make greater use of spatiotemporal information from many frames. Furthermore, the PFRB's multi-channel architecture allows it to perform effectively even with little parameters by employing a type of parameter sharing technique. That created and enhanced the non-local residual block (NLRB) to directly capture long-range spatiotemporal correlations. So, this can be summarized into three major components: a non-local resblock, progressive fusion residual blocks (PFRB), and an upsampling block. The non-local residual blocks are used to extract spatio-temporal characteristics, and PFRB is proposed to fuse them. Finally, the output of a sub-pixel convolutional layer is added to the input frame, which is then upsampled using bicubic interpolation to produce the final super-resolution results.

\section{Our Model and Contribution}
This paper proposes a Generative Adversarial Network that combines the generator of RBPN to achieve high accuracy, and the discriminator of TecoGAN to improve the temporal coherence, with reduced model size.
\subsection{RBPN}
The Recurrent Back Projection network basically calculates the residual images as reconstruction error between the target image and a set of neighboring images, 6 neighboring frames. It exploited temporal relationships between adjacent frames [Haris et al. 2019]. The network mainly consists of three main modules, one feature extraction module, a projection module, and a reconstruction module, as shown in figure 1.
The feature extraction module basically performs two operations, it extracts the features directly from the target frame also extracts the feature from the concatenation of the neighboring frame, and calculates the optical flow from the neighboring frame to the target frame. The second module which is the projection module consists of an encoder and a decoder. The encoder is composed of a multiple image super-resolution (MISR), a single image super-resolution, and residual blocks.
For the decode, it consists of a strided convolution and a residual block. The decoder takes the output of the previous encoder as input to produce the LR features and then fed them to the encoder of the next projection module. The last module, which is the reconstruction module, takes the output of the encoder in each projection module and concatenates them as inputs to produce the final SR results.

The RBPN is chosen specifically as the generator for the proposed network as it contains some modules that jointly use features across layers which are known as the back-projection [Haris et al. 2019]. It offers superior results as it combines the benefits of the original MISR back-projection approach with Deep back-projection networks (DBPNs) which use the back-projection to perform SISR through estimating the SR frame using the LR frame using learning-based models. Comping these two techniques together resulted in a superior accuracy produced by the RBPN network [Irani et al. 1991; Irani et al. 1993].\\

\begin{figure}
\centering
\includegraphics[width=9cm]{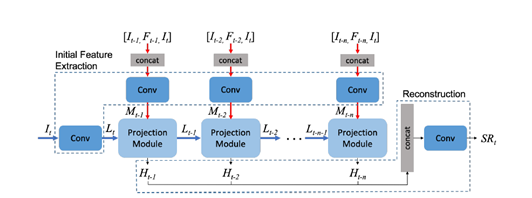}
\caption{RBPN Architecture}
\label{fig: RBPN}
\end{figure}

\subsection{TecoGAN}
In this network, the generator, denoted by G, generates high-resolution (HR) frames from the low-input low-resolution (LR) frames. It takes the LR frames and the previously estimated HR frames as inputs and feeds them into the motion estimation module to get the optical flow. After that, warping is done on the previous HR frames by this optical flow and fed into convolutional modules to get the generated HR frame [Chu et al. 2018]. The discriminator, denoted by D, is Spatio-temporal based and its main role is to compare the generated HR frames with the ground truth.

This network also purposes a new loss function named “Ping-Pong” which focuses on the long-term temporal flow of the generated frames to make the results more natural without artifacts. In addition, it has a relatively low number of parameters for the GAN network about 3 million parameters and hence the inference time is around 42 ms [Chu et al. 2018].
The discriminator guides the generator to learn the correlation between the LR input and the HR targets. In addition, it penalizes G if the generated frames contain less spatial or unrealistic artifacts in comparison to the target HR and the original LR frames. The architecture of the discriminator is shown in figure 2.
\begin{figure}
\centering
\includegraphics[width=6cm]{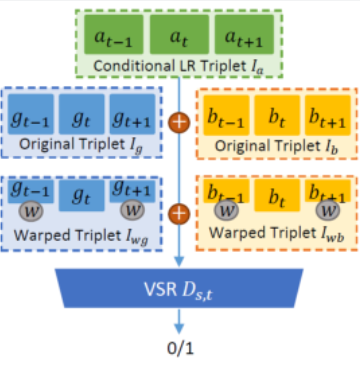}
\caption{Discrimenator Architectue}
\label{fig: TECO}
\end{figure}

\begin{figure*}
\centering
\includegraphics[width=0.9\linewidth]{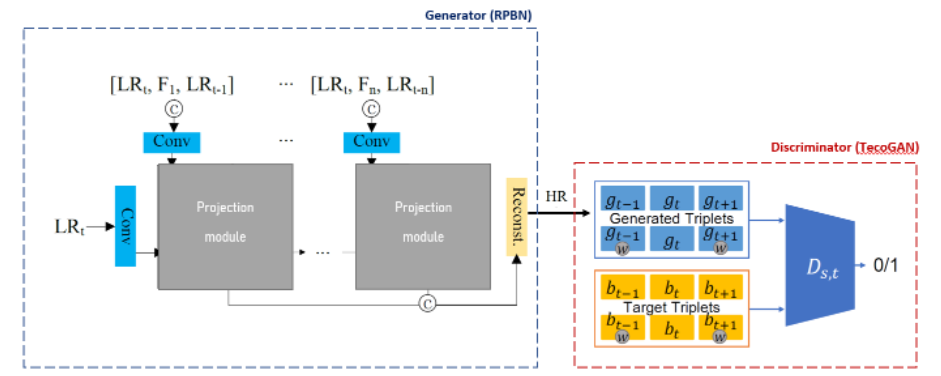}
\caption{RBPGAN Architectue}
\label{fig: RBPGAN}
\end{figure*}

There is an issue that appears when super-resolving at large upscaling factors and that is usually seen with CNNs [Ledig et al. 2017]. So, the proposed network chose TecoGAN as a discriminator to mitigate the issue of a lack of finer texture details. The discriminator is trained to differentiate between super-resolved images and original photo-realistic images.\\

The RBPN is chosen specifically as the generator for the proposed network as it contains some modules that jointly use features across layers which are known as the back-projection [Haris et al. 2019]. It offers superior results as it combines the benefits of the original MISR back-projection approach with Deep back-projection networks (DBPNs) which use the back-projection to perform SISR through estimating the SR frame using the LR frame using learning-based models.

Mainly our proposed architecture, called RBPGAN, combines the virtue of RBPN and TecoGAN as a generator and discriminator respectively. The main goal is to recover precise photo-realistic textures and motion-based scenes from heavily down-sampled videos accordingly this improves the temporal coherence, with reduced model size. The architecture for the proposed architecture is shown in figure 3.

\section{Datasets and Metrics}

\subsection{Datasets}
We used Vimeo-90k and the training dataset created by TecoGAN publishers (we will refer to as VimeoTecoGAN) for training experiments, and we used Vid4 and ToS3 for testing. You can see more details about the datasets in table 1. We produced the equivalent LR frame for each HR input frame when training our model by conducting 4 down-sampling with bicubic interpolation (also known as Gaussian Blur method). Thus, we achieve self-supervised learning by producing the input-output pairs for training automatically, without any human interaction. For testing purposes, we obtained comparable assessment findings on the Tears of Steel data-sets (room, bridge, and face - referred to as ToS3 scenes) in addition to the previously analyzed Vid4 dataset. To make the outputs of all methods comparable, we followed the procedures of previous work [Jo et al. 2018; Sajjadi et al. 2018]. For all result images, we first excluded spatial borders with a distance of 8 pixels to the image sides, then further shrank borders such that the LR input image is divisible by 8, and for spatial metrics, we disregard the first two and last two frames for spatial metrics, while we ignore the first three and last two frames for temporal metrics since one additional prior frame is necessary for inference. Ultimately, we began experimenting with some of our own LR video sessions in which we considered bodily motions. When we compared them to the other datasets and metric breakdowns, we discovered that our measures accurately captured human time perception.

\begin{center}
\includegraphics[width=8cm]{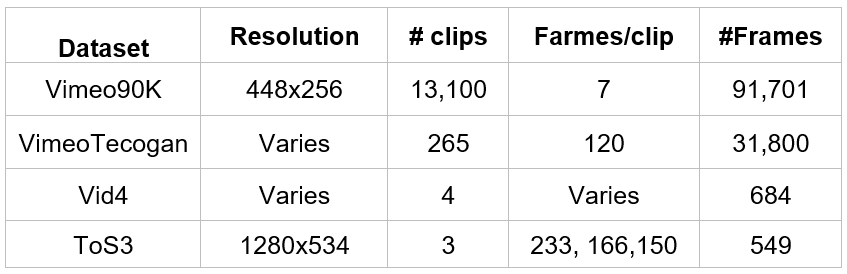}
Table 1: Datasets
\end{center}

\subsection{Evaluation Metrics}
While the visual results offer a first indication of the quality of our technique, quantitative assessments are critical for automated evaluations over greater numbers of samples. Because ground-truth data is available, we will concentrate on the VSR assignment in this section. We give metrics evaluations of several models in relation to existing geographical metrics. We also justify and suggest two new temporal metrics for measuring temporal coherence. The usual criterion for evaluating the quality of super-resolution results mainly includes Peak signal-to-noise ratio (PSNR) and Structural index similarity (SSIM). PSNR is the ratio of an image's maximum achievable power to the power of corrupting noise that affects the quality of its representation. To calculate the PSNR of a picture, it must be compared to an ideal clean image with the highest potential power. Higher outcomes are preferable. A single SR frame's PSNR can be calculated as
$$
PSNR = 10l( \frac{MAX^2}{MSE} ) \eqno{(1)}
$$
where MAX is the color value's maximum range, which is commonly 255 and MSE is the mean squared error. Generally, a greater PSNR value indicates higher quality. While SSIM measures the similarity of structure between two corresponding frames using an uncompressed or distortion-free image as a baseline. A higher SSIM value indicates higher quality. PSNR may be more sensitive to Gaussian noise, whereas SSIM may be more sensitive to compression errors. Their values, however, are incapable of reflecting video quality for human vision. That implies, even if a video has a very high PSNR value, it may still be unpleasant for humans. As a result, deep feature map-based measures like LPIPS [Zhang et al. 2018] can capture more semantic similarities. The distance between picture patches is calculated using LPIPS (Learned perceptual image patch similarity). Higher implies more distinct. Lower values indicate a closer match.LPIPS indicates the perceptual and semantic similarity to the ground truth. In other words, lower LPIPS means a more natural video. Additionally, ToF is used to calculate the pixel-wise difference of movements inferred from successive frames.

$$
tOF = ||OF(b_{t-1}, b_t ) - OF(g_{t-1}, g_t )|| \eqno{(2)}
$$

\section{Loss Functions}

The loss functions used while training our model are as follows:
\begin{enumerate}
    \item GAN Loss (min-max loss):\\
    We use the Vanilla GAN loss, which is the simplest form of the GAN loss, for the adversarial training. The generator tries to minimize it while the discriminator tries to maximize it. \\
    \begin{center}
    \includegraphics[width=7cm]{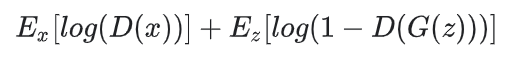}
    Eqn.1
    \end{center}

    Here, $D(x)$ is the discriminator's estimate of the probability that real data instance x is real, and $D(G(z))$ is the discriminator's estimate of the probability that a fake instance is real. E is the expected value over all data instances.\\
    
    \item Pixel loss:\\
    Minimizes the pixel-wise squared differences between Ground Truth and generated frames.\\
    \begin{center}
    \includegraphics[width=2.5cm]{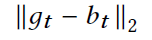} 
    Eqn.2
    \end{center}

    \item Ping Pong Loss:\\
    Proposed by TecoGAN model, effectively avoids the temporal accumulation of artifacts, and targets generating natural videos that are consistent over time. PP loss uses a sequence of frames with the forward order as well as its reverse. Using an input number of frames of length n, we can form a symmetric sequence $a_1, ...a_{n-1}, a_n, a_{n-1}, ... a_1$ such that when feeding it to the generator, the forward results should be identical to the backward result [Chu et al. 2018].\\
    
    \begin{center}
    \includegraphics[width=3cm]{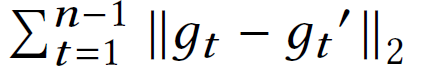}
    Eqn.3
    \end{center}

    Here, the forward results are represented with $g_t$ and the backward results with ${g_t}'$\\
    
    \item Feature/perceptual Loss:\\
    Encourages the generator to produce features similar to the ground truth ones by increasing the cosine similarity of their feature maps. It ensures more perceptually realistic and natural generated videos.
    Our discriminator features contain both spatial and temporal information and hence are especially well suited for the perceptual loss.\\
    
    \begin{center}
    \includegraphics[width=4cm]{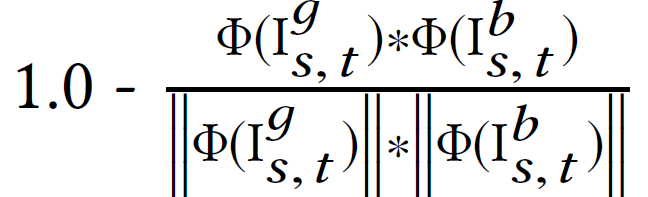} 
    Eqn.4
    \end{center}

    Where ${I^g} = \{{g_{t-1},g_t,g_{t+1}}\}, I^b = \{{b_{t-1},b_t,b_{t+1}}\}$\\
    
    \item Warping Loss:\\
    Used while training the motion estimation network (F) that produces the optical flow between consecutive frames.\\
    
    \begin{center}
    \includegraphics[width=5cm]{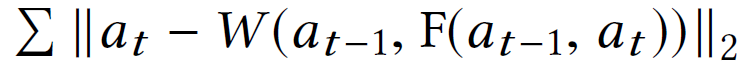} 
    Eqn.5
    \end{center}

    Where $W()$ is the warping function, $F()$ is the flow estimator, and $a_t$ is the LR frame in position $t$.

\end{enumerate}

\section{Experiments}

During the training process, GANs' generative and discriminative models can play games with each other to achieve greater perceptual quality than other standard models. As a result, GANs are widely used in the field of Super Resolution. To deal with large-scale and unknown degradation difficulties in VSR tasks, we depend on the remarkable ability of GAN models' deep feature learning. We also refer to the TecoGAN method's design and introduce the Spatio-temporal adversarial structure to aid the discriminator's understanding and learning of the distribution of Spatio-temporal information, avoiding the instability impact in the temporal domain that standard GANs suffer from. We also introduced a more accurate generator module based on the RBPN model, into the TecoGAN design to ensure the quality and improve the temporal coherence.\\

In all our experiments, we focus on the 4× Super Resolution factor since it provides satisfactory results and takes a reasonable amount of training. Also, we used crop size of 32x32 and gaussian downsampling. All experiments were done using following specifications to enable the dense nature of the training phase: 64GB of DDR4 RAM, 2.80GHz Intel Core i9-10900F CPU, NIVIDIA GetForce RTX 3090 (1 x 24 GB) GPU, and Ubuntu 20.04.3 LTS operating system.\\

We will now present and explain the experiments we did in sequence, and later we will explain and discuss their results comparatively.\\

So, we started by training and testing our two base models (TecoGAN and RBPN) to ensure their correctness and reliability before integrating them and produce our model. Then, we integrated them as discussed in section 2. Later, we performed some experiments on our model with different parameters, loss functions, etc. till we reached the best outcome. The final model is later compared with the other state-of-the-art models in terms of PSNR, SSIM, LPIPS, and ToF metrics.\\

\begin{enumerate}

    \item Experiment 1: Reduced RBPN Model Size\\ 
    As discussed, RBPN is the base model we are using for our model's generator, and we started by training and testing it. The model size was very large, and we encountered memory-related issues and therefore had to reduce its size. So, we decreased the number of neighbour frames passed to the projection modules in the model, which resulted in a decreased size and solving our problems. The training of this experiment took around 1 hour/epoch and we trained it for 150 epochs, using VimeoTecoGAN dataset, and other parameters as the original published model.
    
    \item Experiment 2: Lightweight TecoGAN\\
    We trained and tested the TecoGAN model, which demonstrated very adequate results with a fewer number of parameters in comparison with the rest of the state-of-the-art models. We utilized 1 GPU for the first training experiment of TecoGAN implemented using the TensoFlow framework. The results were very encouraging; however, the training took more than 170 hours to complete. So the structure of the network was kept the same, but implemented to be more lighwight and less computationally dense. The new implementation provides a model with a smaller size and better performance than the official TecoGAN implementation as you can notice from table(2). We also trained it using more computationally powerful machine (with 24GB GPU), and the results show that the reduced model has less training time than the official implementation by a factor of 6.7x. Also, this implementation is done using PyTorch framework to make it compatible with RBPN in the integration phase later.

    \begin{center}
    \includegraphics[width=8cm]{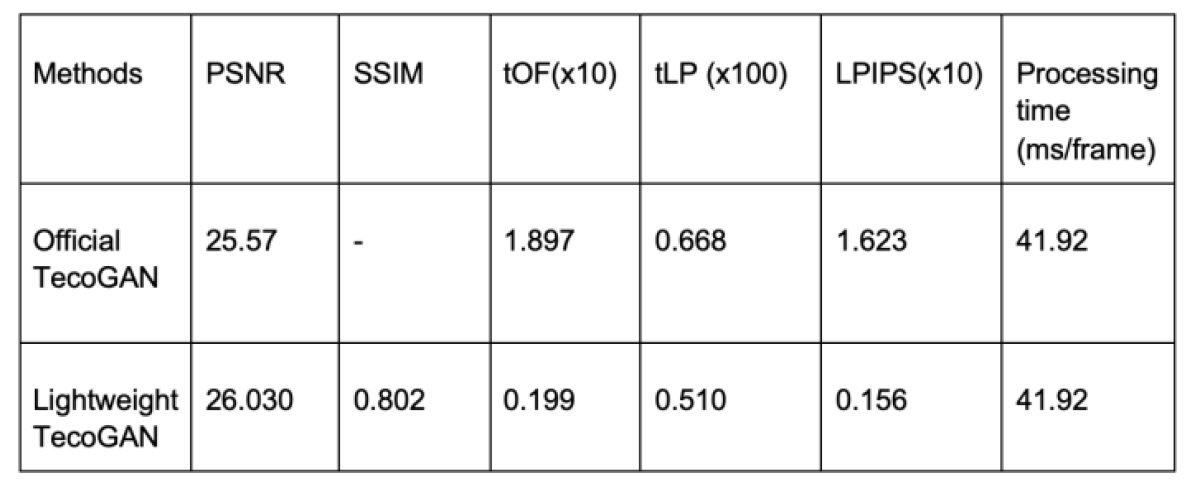} 
    Table 2: Comparison between Official TecoGAN implementation and reduced model performance.
    \end{center}

    \item Experiment 3: Model Integration\\
    After we ensured the correctness, reliability and readiness of the two base models for the integration phase, we began integrating RBPN as the generator with the spatio-temporal discriminator from TecoGAN to create our GAN model and prepare it for some experiments. The integration was a challenging task since the two models had many differences in functions' interfaces, dependencies used, training datasets and the lack of generalization to fit any other dataset, and the coding style. We experimented in two ways: Replacing the existing generator of TecoGAN with RBPN in TecoGAN environment, and the second is addng the Spatio-Temporal discriminator of TecoGAN to RBPN and transforing a feed-forward model to a generative model. Lastly, after solving all issues, we produced our model: RBPGAN-Recurrent Back Projection Generative Adversarial Network.

\end{enumerate}

We will now discuss the experiments done on RBPGAN (our model) to monitor the model potential and test our hypothesis. 

\begin{enumerate}
    \item Experiment 4.1: \\
    We used all loss functions mentioned in the previous section (Ping Pong loss, Pixel loss, Feature loss, Warping loss, and GAN loss). We used 2 neighbour (adjacent) frames per frame, but due to the use of ping pong loss, this number is doubled to create the backward and forward paths. Therefore, the generator was using 4 neighbour frames per frame. We trained both the generator and discriminator simultaneously from the beginning, using the VimeoTecoGAN dataset, and training took around 3.5 days using the specs mentioned. 
    \item Experiment 4.2: \\
    We used the same loss functions as experiment 4.1 except the ping pong loss to observe its effect on the results. We used 3 neighbour frames per frame, started the training of generator and discriminator together, and used the same dataset and other parameters as experiment 4.1. The training took around 3 days.
    \item Experiment 4.3:\\
    This experiment is the same as experiment 4.2 except that we firstly trained the generator solely for some epochs and then started the training of the GAN using this pre-trained part. The training took aound 3 days using the same dataset, number of neighbours and other parameters.
    \item Experiment 5:\\
    We trained RBPN model with the same number of neighbours, crop size, dataset, and other unify-able parameters as we did for our model in the 3 previous experiments to ensure fair comparison between it and our model.
\end{enumerate}
\pagebreak

\section{Results}

\begin{center}

Following are the results and metrics evaluation for the experiments done and explained in previous section (table 3,4):
\\~\\

\begin{tabular}{ |p{1.5cm}|p{1.5cm}|p{1.5cm}|p{1.5cm}|  }
 \hline
 Metric Name&Experiment 4.1 &Experiment 4.2&Experiment 4.3\\
 \hline
 PSNR   & 25.58    &\textbf{25.74}&   25.56\\
 LPIPS&   1.47  & \textbf{1.44}   &1.45\\
 tOF &2.46 & \textbf{2.35}&  2.40\\
 SSIM    &0.756 & \textbf{0.762}&  0.751\\
 \hline
\end{tabular}\\
Table 3: Comparative analysis between all the conducted experiments on our model for \textbf{Vid4} dataset

\begin{tabular}{ |p{1.5cm}|p{1.5cm}|p{1.5cm}|p{1.5cm}|  }
 \hline
 Metric Name&Experiment 4.1 &Experiment 4.2&Experiment 4.3\\
 \hline
 PSNR   & \textbf{32.89}    &32.85&   32.78\\
 LPIPS&   0.78  & \textbf{0.69}   &0.75\\
 tOF &\textbf{1.60 }& 1.64&  1.62\\
 SSIM    &0.872 & \textbf{0.880}&  0.869\\
 \hline
\end{tabular}
\\~\\
Table 4: Comparative analysis between all the conducted experiments on our model for \textbf{ToS3} dataset \\
\end{center}

So, overall, experiment 4.2 yields the best results collectively, and therefore we will use it to compare with the state-of-the-art models (tables 5,6). Also, you see in figures 4 and 5 some examples from vid4 dataset for our model. 

\begin{figure}
\centering
\includegraphics[width=7cm]{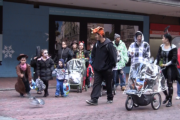}
\includegraphics[width=7cm]{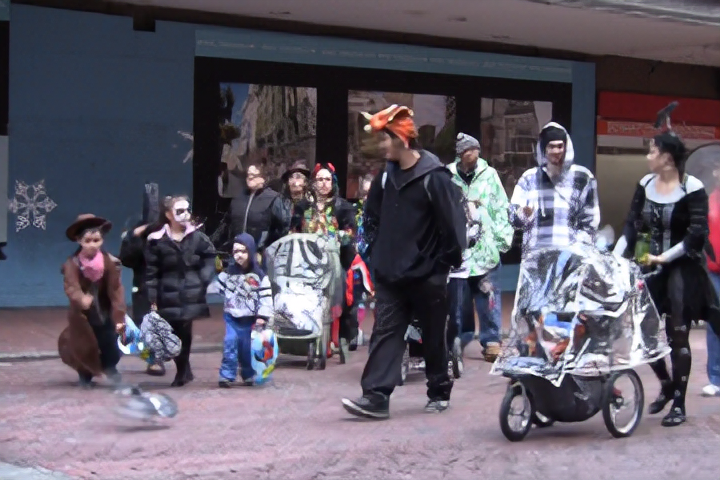}
\caption{walk (Top:LR , Bottom: RBPGAN)}
\label{fig: walk lr}
\end{figure}

\begin{figure}
\centering
\includegraphics[width=6cm]{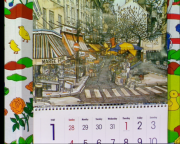}
\includegraphics[width=6cm]{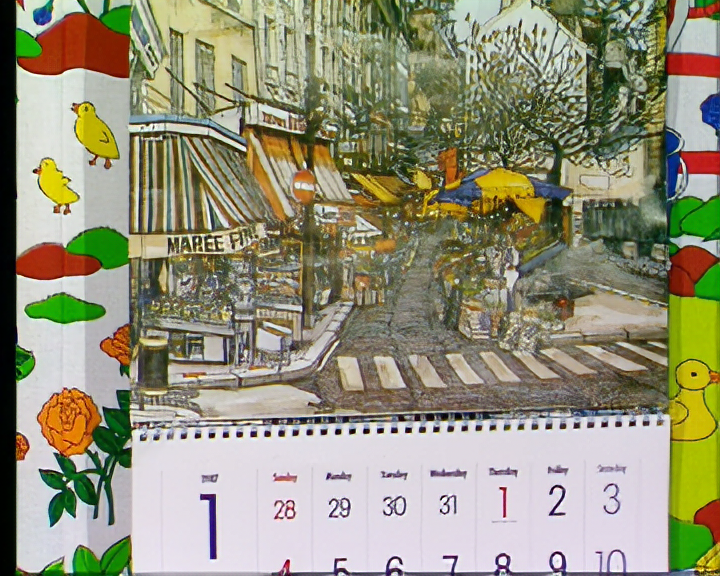}
\caption{calendar (Top:LR , Bottom: RBPGAN)}
\label{fig: calendar lr}
\end{figure}

\newpage

\begin{table*}
  \centering
    \label{tab:widetable}
\begin{tabular}{ |p{2cm}|p{2cm}|p{2cm}|p{2cm}| p{2cm} |p{2cm}| p{2cm} | }
 
 \hline
 Metric &Experiment 4.2 (Ours) &TecoGAN&RBPN (3 neighbors) & BIC & ENet & DuF\\
 \hline
 PSNR   & 25.74    &25.57&   26.71 & 23.66 & 22.31 & \textbf{27.38}\\
 LPIPS&   \textbf{1.44}  & 1.62   &2.0  & 5.04 & 2.46 & 2.61\\
 tOF &2.35 & 1.90&  2.19 & 5.58 & 4.01  & \textbf{1.59}\\
 SSIM    &\textbf{0.756} & 0.770 &  0.801 & NA & NA & 0.815\\
 \hline
\end{tabular}

Table 5: Comparison between experiment 2 and state-of-the-arts  for \textbf{Vid4} dataset
\end{table*}

\begin{table*}
  \centering
    \label{tab:widetable}

\begin{tabular}{ |p{2cm}|p{2cm}|p{2cm}|p{2cm}| p{2cm} |p{2cm} |p{2cm}  |}

 \hline
 Metric&Experiment 4.2 (Ours) &TecoGAN&RBPN (3 neighbors) & BIC & ENet & DuF\\
 \hline
 PSNR   & 32.85    &32.65& 34.32 & 29.58 & 27.82 & \textbf{34.6}\\
 LPIPS&   \textbf{0.69}  & 1.09 & 1.10 & 4.17 & 2.40 & 1.41\\
 tOF &1.64 & 1.34&  1.54 & 4.11 & 2.85 & \textbf{1.11}\\
 SSIM    &\textbf{0.880} &0.892 &  0.915 & NA & NA & NA\\
 \hline
\end{tabular}

Table 6: Comparison between experiment 2 and state-of-the-arts  for \textbf{ToS3} dataset
\end{table*}

\section{Discussion}

As per our hypothesis, we were trying to merge the highly realistic output of RBPN with the temporally coherent output of TecoGAN to have a smooth high quality output for our model. In the results section, we showed the metrics of output for our model and how we achieve higher qualities when compared with TecoGAN and higher temporal cohesion when compared with RBPN. However, we were hoping our model would generate HR videos with qualities equivalent to RBPN, yet the quality of the generated were not less than that of RBPN, although higher than TecoGAN. This is due to the GAN training we added to our model and the huge need for fine-tuning. Due to time and hardware limitations, we decided on training all models for the same number of epochs although some models used different datasets and GANs typically need more time to achieve convergence. We only trained our GAN model for 51 epochs although the model might have achieved higher stability at a higher number of epochs.

When assessing our model with the LPIPS metric, which is responsible for measuring the temporal cohesion, we found that our model surpasses both base models, TecoGAN and RBPN. This is because the discriminator helps the generator in learning further temporal cohesion. Further improvements for other metrics could have been reached if we had more time for fine-tuning our GAN model. We could have tried different combinations of loss functions and different learning rates till we reach the optimal training conditions.

\section{Limitations and Further work}

Our model needed more powerful computational resources but we were only limited to a 2-GPU machine with a 64GB memory which limited our experiments since each experiment took longer than usual on the available machines. Moreover, while our technique produces extremely realistic results for a large variety of natural pictures, in some circumstances, such as under-resolved faces and text in VSR, or jobs with dramatically differing motion between two domains, our method can provide temporally coherent but sub-optimal details. It would be interesting to use both our technique and motion translation from contemporaneous work in the latter instance [Chen et al. 2019]. Hence, we recommend using different downsampling methods to introduce more generalization to the model, and may be train the model on more augmented datasets that focus more on faces and text.

Although RBPGAN combines temporal coherence and high video accuracy, couple of ideas could be investigated to improve it. Because it often comprises topics such as individuals, the foreground receives far more attention than the backdrop in visual images. To increase perceptual quality, we may separate the foreground and background and have RBPGAN execute "adaptive VSR" by using different rules for each. For example, we may use a larger number of frames to extract features from the foreground compared to the backdrop. In addition, there is continuing research on the usage of accelerator techniques to boost the speed of network training and inference time, leading to real-time VSR transition. The most relatable ones we found are; Convolutional Computation Acceleration, Efficient Upsampling, and Batch Normalization Fusion we anticipate that they will provide a very useful basis for a wide range of generative models for real-time HR video generation.

\section{conclusion}
We managed to approve our hypothesis and achieved the highest results when it comes to temporal coherence. While current adversarial training produces generative models in a variety of fields, temporal correlations in the generated data have received far less attention. We, on the other hand, concentrate on improving learning goals and present a temporally self-supervised approach to bridge this gap. For the sequential generation tasks such as video super-resolution and unpaired video translation, natural temporal shifts are critical. The reduced Recurrent Back-Projection Network in our model extracts information from each context frame and then combines and processes them within a refinement framework based on the back-projection idea in multiple-frame super resolution. The inter-frame motion is estimated with respect the the target, which also aids in producing more temporally coherent videos.

\section{acknowledgement}
We wish to acknowledge the help provided by the technical and support staff in the Computer Science and Engineering department of AUC the American University in Cairo.
We would also like to show our deepest appreciation to our supervisors Prof. Dr.Cherif Salama, Prof. Dr.Hesham Eraqi,and Prof.Dr. Moustafa Youssef who guided us through the project.

\bibliographystyle{IEEEtran}  
\bibliography{IEEEexample}

[1] A. Lucas, S. LApez-Tapia, R. Molina, and A. K. Katsaggelos, “Generative adversarial networks and perceptual losses for video super-resolution,” IEEE Trans. Image Process., vol. 28, no. 7, pp. 3312– 3327, July 2019.\\~\\

[2] Adji B Dieng, Yoon Kim, Alexander M Rush, and David M Blei. Avoiding latent variable collapse with generative skip models. arXiv preprint arXiv:1807.04863, 2018.\\~\\

[3] Cao, Y., Wang, C., Song, C., Tang, Y., and Li, H. (2021, July). Real-Time Super-Resolution System of 4K-Video Based on Deep Learning. In 2021 IEEE 32nd International Conference on Application-specific Systems, Architectures and Processors (ASAP) (pp. 69-76). IEEE.\\~\\

[4] Chu, M., Xie, Y., Leal-Taixé, L., and Thuerey, N. (2018). Temporally coherent gans for video super-resolution (tecogan). arXiv preprint arXiv:1811.09393, 1(2), 3.\\~\\

[5] Haris, M., Shakhnarovich, G., and Ukita, N. (2019). Recurrent back-projection network for video super-resolution. In Proceedings of the IEEE/CVF Conference on Computer Vision and Pattern Recognition (pp. 3897-3906).\\~\\

[6] Irani, M.; Peleg, S. Improving resolution by image registration. CVGIP: Graphical Models and Image Processing Vol. 53, No. 3, 231–239, 1991.\\~\\

[7] Irani, M.; Peleg, S. Motion analysis for image enhancement: Resolution, occlusion, and transparency. Journal of Visual Communication and Image Representation Vol. 4, No. 4, 324–335, 1993.\\~\\

[8] Ledig, C.; Theis, L.; Huszar, F.; Caballero, J.; Cunningham, A.; Acosta, A.; Aitken, A.; Tejani, A.; Totz, J.; Wang, Z. et al. Photo-realistic single image super-resolution using a generative adversarial network. In: Proceedings of the IEEE Conference on Computer Vision and Pattern Recognition, 4681–4690, 2017.\\~\\

[9] M. Chu, Y. Xie, J. Mayer, L. Leal-Taix´e, and N. Thuerey, “Learning Temporal Coherence via Self-Supervision for GAN-based Video Generation,” ACM Transactions on Graphics,2020\\~\\

[10] M. Haris, G. Shakhnarovich, and N. Ukita, “Recurrent back-projection network for video superresolution,” in Proc. IEEE Conf. Comput. Vis. Pattern Recognit., 2019, pp. 3892–3901.\\~\\

[11] P. Yi, Z. Wang, K. Jiang, J. Jiang, and J. Ma, “Pro- gressive fusion video super-resolution network via exploiting non-local spatio-temporal correlations,” in Proc IEEE Int. Conf. Comput. Vis., 2019 , pp. 3106– 3115.\\~\\

[12] X. Wang, K. C. K. Chan, K. Yu, C. Dong, and C. C. Loy, “EDVR: Video restoration with enhanced deformable convolutional networks,” in Proc. IEEE Conf. Comput. Vis. Pattern Recognit. Workshops, 2019, pp. 1954–1963.\\~\\

[13] Y. Jo, S. W. Oh, J. Kang, and S. J. Kim, “Deep video super-resolution network using dynamic upsampling filters without explicit motion compensation,” in Proc. IEEE Conf. Comput. Vis. Pattern Recognit., 2018, pp. 3224–3232\\~\\

[10] M. Haris, G. Shakhnarovich, and N. Ukita, “Recurrent back-projection network for video superresolution,” in Proc. IEEE Conf. Comput. Vis. Pattern Recognit., 2019, pp. 3892–3901.\\~\\

[11] P. Yi, Z. Wang, K. Jiang, J. Jiang, and J. Ma, “Pro- gressive fusion video super-resolution network via exploiting non-local spatio-temporal correlations,” in Proc IEEE Int. Conf. Comput. Vis., 2019 , pp. 3106– 3115.\\~\\

[12] X. Wang, K. C. K. Chan, K. Yu, C. Dong, and C. C. Loy, “EDVR: Video restoration with enhanced deformable convolutional networks,” in Proc. IEEE Conf. Comput. Vis. Pattern Recognit. Workshops, 2019, pp. 1954–1963.\\~\\

[13] Y. Jo, S. W. Oh, J. Kang, and S. J. Kim, “Deep video super-resolution network using dynamic upsampling filters without explicit motion compensation,” in Proc. IEEE Conf. Comput. Vis. Pattern Recognit., 2018, pp. 3224–3232.\\~\\

\addtolength{\textheight}{-12cm}  

\end{document}